\documentclass[letterpaper, 10 pt, conference]{ieeeconf}
\usepackage{PackageHeading}
\usepackage[absolute,overlay]{textpos}

\usepackage{booktabs}
\usepackage{siunitx}
\usepackage{caption}

\makeatletter
\let\labelindent\relax  
\makeatother
\usepackage[shortlabels]{enumitem}  

\usepackage{enumitem}
\usepackage{array}    

\usepackage{fancyhdr}
\usepackage[hidelinks]{hyperref}
\usepackage{url}
\fancypagestyle{firstpage}{
  \fancyhf{}

  \fancyfoot[C]{\footnotesize
    \parbox{\dimexpr\textwidth-2em\relax}{%
      ©~2024 IEEE. Personal use of this material is permitted. Permission from IEEE must be obtained for all other uses, in any current or future media, including reprinting/republishing this material for advertising or promotional purposes, creating new collective works, for resale or redistribution to servers or lists, or reuse of any copyrighted component of this work in other works.
    This is the authors’ accepted manuscript for IROS 2024. The final version of record is available at:
    \href{https://doi.org/10.1109/IROS58592.2024.10802717}{10.1109/IROS58592.2024.10802717}.}}
}

\usepackage{tikz}
\usetikzlibrary{positioning}
\usepackage{svg}

\usepackage{makecell} 

\title{\Large \bf
Reinforcement Learning of Dolly-In Filming Using a Ground-Based Robot
}

\author{Philip Lorimer$^{1}$, Jack Saunders$^{1}$, Alan Hunter$^{2}$, and Wenbin Li$^{1}$
\thanks{This work was supported by EPSRC Centre for Digital Entertainment with the grant number EP/L016540/1.}
\thanks{$^{1}$Department of Computer Science,
        University of Bath, UK, \{\protect\url{pall20, js3442, w.li}\}\protect\url{@bath.ac.uk}}
\thanks{
        $^{2}$Department of Mechanical Engineering,
        University of Bath, UK, \protect\url{A.J.Hunter@bath.ac.uk}}%
}

\begin{document}

\maketitle
\thispagestyle{firstpage}
\pagestyle{empty}

\begin{abstract}
Free-roaming dollies enhance filmmaking with dynamic movement, but challenges in automated camera control remain unresolved. Our study advances this field by applying Reinforcement Learning (RL) to automate dolly-in shots using free-roaming ground-based filming robots, overcoming traditional control hurdles. We demonstrate the effectiveness of combined control for precise film tasks by comparing it to independent control strategies. Our robust RL pipeline surpasses traditional Proportional-Derivative controller performance in simulation and proves its efficacy in real-world tests on a modified ROSBot 2.0 platform equipped with a camera turret. This validates our approach's practicality and sets the stage for further research in complex filming scenarios, contributing significantly to the fusion of technology with cinematic creativity. This work presents a leap forward in the field and opens new avenues for research and development, effectively bridging the gap between technological advancement and creative filmmaking.
\end{abstract}


\section{Introduction}
Robotic technology has dramatically transformed the filmmaking landscape, offering unprecedented flexibility in camera movement and expanding creative possibilities \cite{BogueRobert2022Tror}. Free-roaming dolly systems have evolved from traditional, track-constrained designs, enabling more flexible, three-dimensional camera movement. This technological advancement, however, introduces significant challenges in control adaptability and operational consistency, particularly in executing traditional film techniques like dolly-in shots. The reliance on operators with varying skills exacerbates these challenges, introducing variability in film quality that can detract from the system's potential. While some classical control methods have been applied in similar domains, our research uses Reinforcement Learning (RL) to develop an advanced control mechanism that automates these techniques, providing a foundation for more precise and adaptable robotic cinematography.

Within this context, RL \cite{sutton2018reinforcement} emerges as a promising avenue to enhance the adaptability and precision of ground-based robotic filming systems. Prior studies have extensively explored the capabilities of autonomous systems in filmmaking, with particular emphasis on aerial cinematography \cite{Goh2021AerialFW} and static camera work. These investigations highlight the potential of computer vision and classical control methods for task planning and execution, demonstrating notable successes in their application \cite{Chen2014, 10.1145/2502081.2502086, inproceedings}. Concurrently, the shift towards data-driven approaches, particularly RL, offers significant adaptability by learning from data and reducing dependence on explicit models \cite{Chen2014, Chen2013}. Specifically, RL has shown considerable promise in mobile aerial filming and other applications, excelling at complex objectives such as framing and executing shots on diverse platforms \cite{10.1007/s10462-021-09997-9, doi:10.1177/17298814211007305, 2019arXiv190402579G, passalis_deep_2019}. Despite these advancements, a notable gap remains in applying such innovations to ground-based robotic filming, particularly in managing complex coordination and control challenges.

Incorporating RL for controlling ground-based robots offers a pathway to execute dolly-in shots with greater versatility. In this work, 'Independent' control refers to the separate handling of actions such as throttle and steering. In contrast, 'Combined' control refers to the unified management of these actions within a single system. The distinction is critical, as the choice of these control architectures significantly impacts platform movement and camera framing. Aerial cinematography research often prefers independent control for its flexibility in meeting specific component needs \cite{2019arXiv190402579G, Leottau2016DecentralizedRL}. However, managing these controls separately can be challenging, particularly when integrating control disturbances, which can impair coordination and performance without sophisticated disturbance modelling \cite{e24111681}. In contrast, combined RL control treats the system as a unified entity, streamlining management and enhancing action coordination \cite{Hayes_2022}, offering a more practical approach for complex filming tasks.

Given these themes, our investigation addresses this gap by leveraging RL to autonomously control a ground-based robot in executing the traditional dolly-in shot, a staple in cinematic production. Central to our methodology is the abstraction of the robot and task dynamics. By distilling these into essential elements, we refine our RL approach to emphasise core cinematic movements, creating a scalable framework that accommodates the complexities of robotic cinematography. This abstraction showcases the effectiveness of RL within known contexts and supports extending our approach to tackle more intricate filming tasks and diverse environmental conditions.

The main contributions of this work are as follows:

\begin{itemize}[noitemsep] 

    \item We evaluate combined and independent control strategies for dolly-in shots through throttle and steering actions on a ground-based filming platform, highlighting performance differences and their applicability in filmmaking.
    
    \item  The direct application of a combined control RL agent to a ground-based filming robot achieves dolly-in shot performance matching that of a finely tuned PD controller, as demonstrated across simulated and real-world tests.

    \item We demonstrate a pipeline to train RL agents on ground-based filming robots, enabling them to simulate and execute successful dolly-in shots in real-world tests, achieving a strong positive correlation in zero-shot Sim2Real transfers.

\end{itemize}

This work addresses the challenges posed by advanced robotic technologies in filmmaking, highlighting the untapped potential of reinforcement learning (RL) in this field. By framing our contributions within the broader discourse on autonomous camera systems, we illuminate a path toward achieving greater creative flexibility and technical precision in cinematography. Our findings lay a robust foundation for future research to leverage RL for more complex and creatively demanding tasks on mobile ground-based filming robots. This research not only pushes the boundaries of robotic cinematography but also establishes an abstraction approach as the backbone of our methodology, ensuring that our solutions are effective, adaptable, and scalable, thus setting the stage for continued advancements in this domain.

The paper proceeds as follows: Section \ref{sec:formalisation_of_dolly_in} formalises the dolly-in shot. Section \ref{sec:RL_agents} covers reinforcement learning techniques and agent configurations, while  \ref{sec:Experimental_Results} details our experiments and results. Discussions and conclusions are in sections \ref{sec:discussion_lims} and \ref{sec:conclusions}.
  
\begin{figure*}[ht]
\centering
\includegraphics[width=0.24\linewidth]{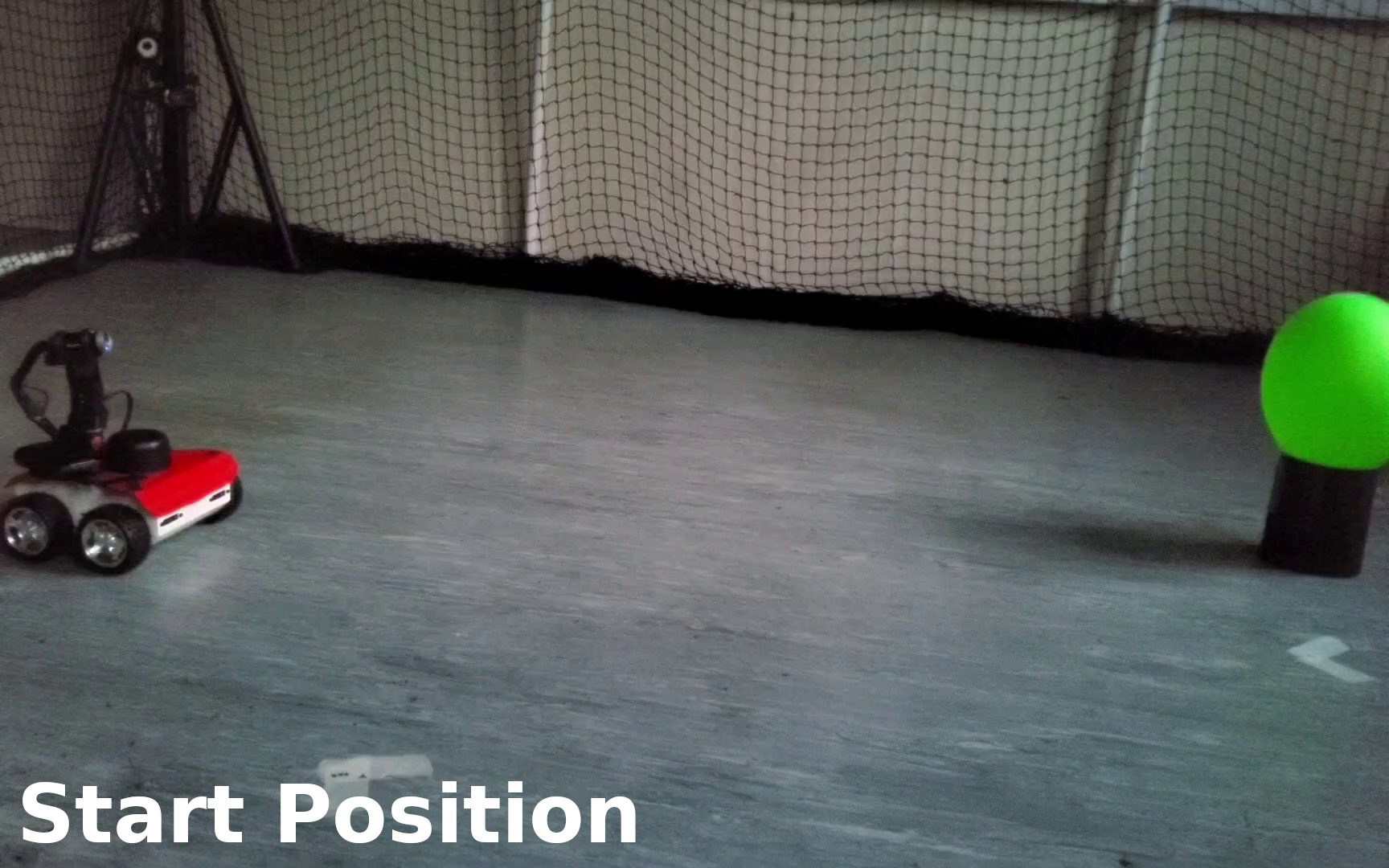}
\includegraphics[width=0.24\linewidth]{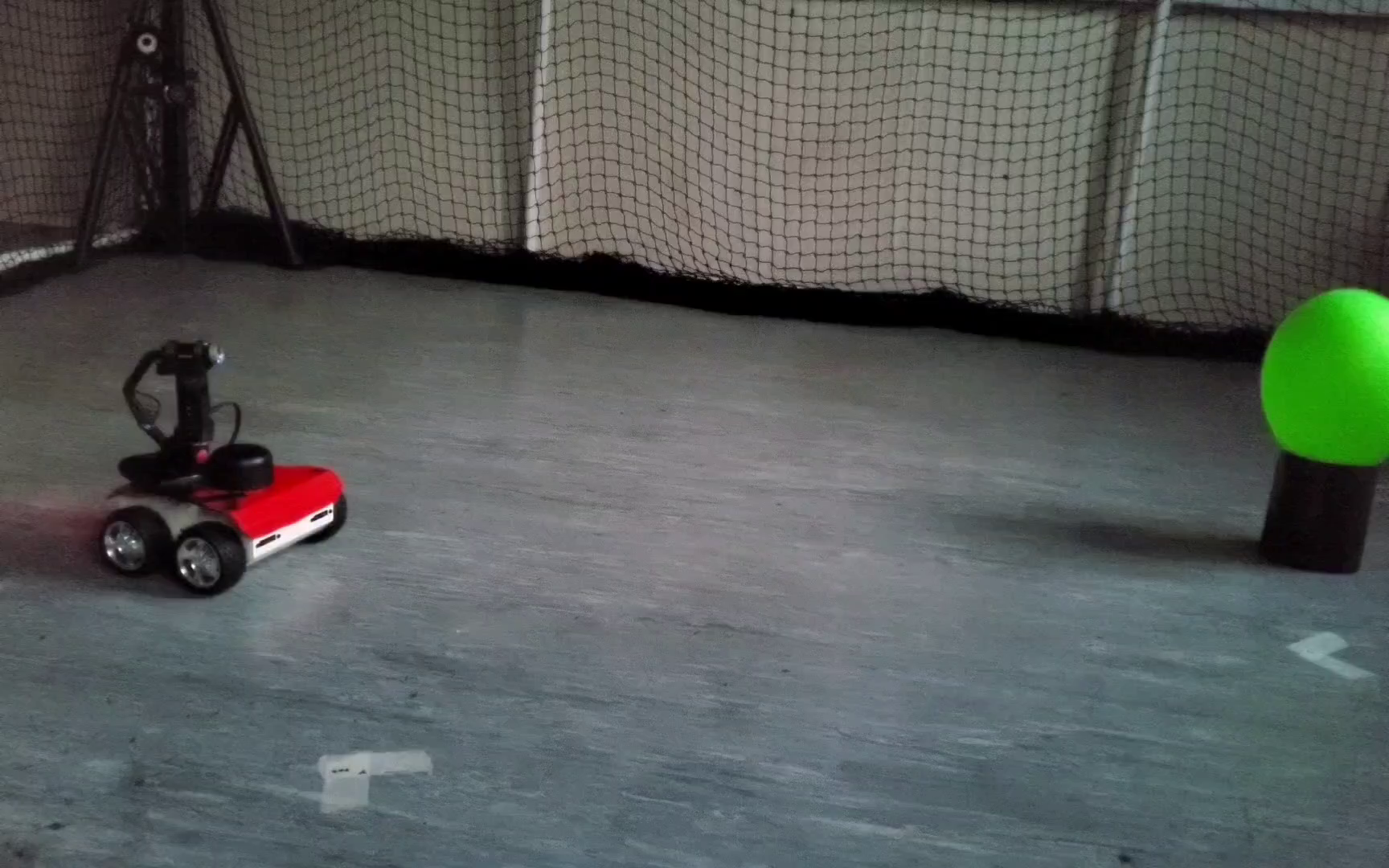}
\includegraphics[width=0.24\linewidth]{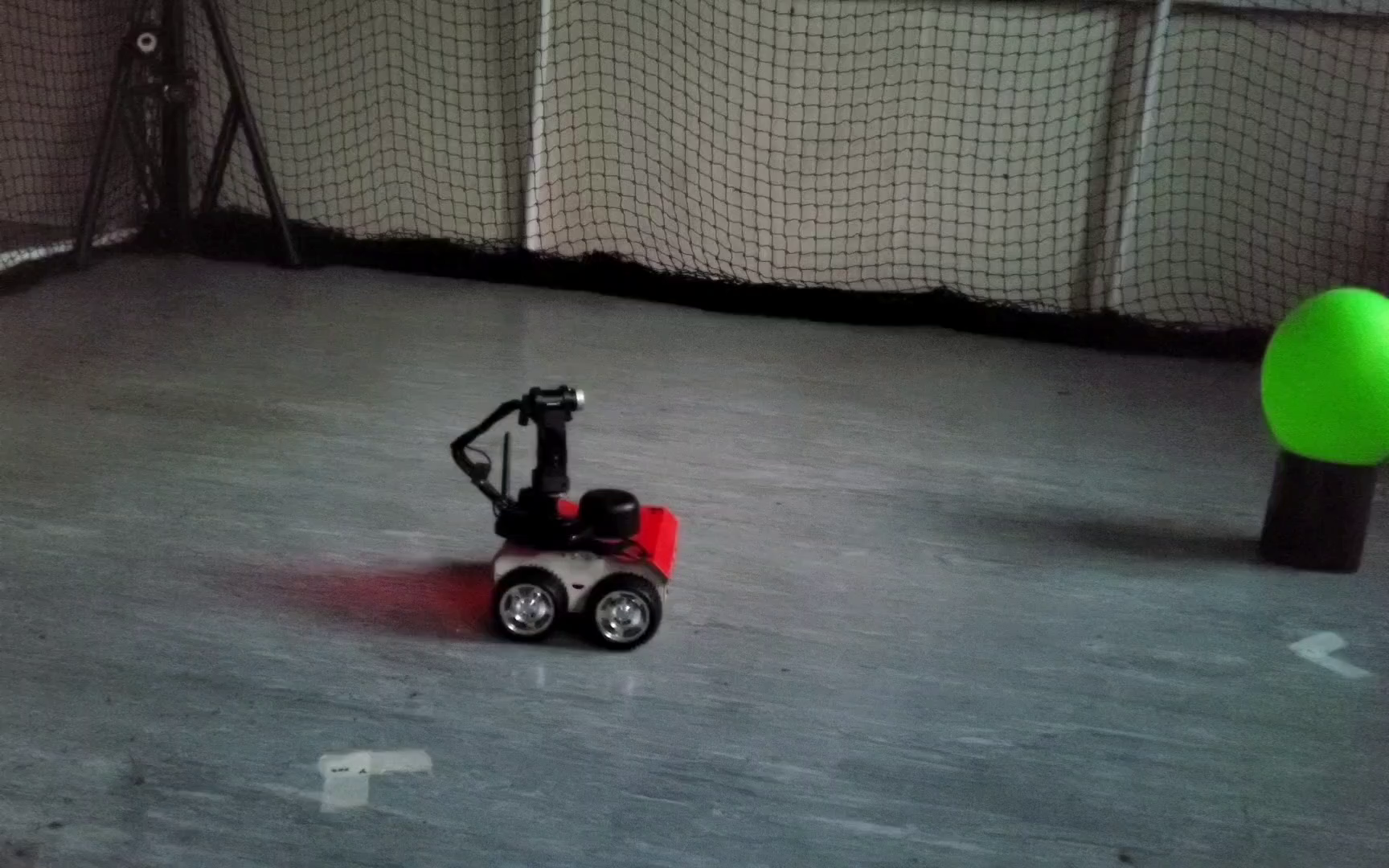}
\includegraphics[width=0.24\linewidth]{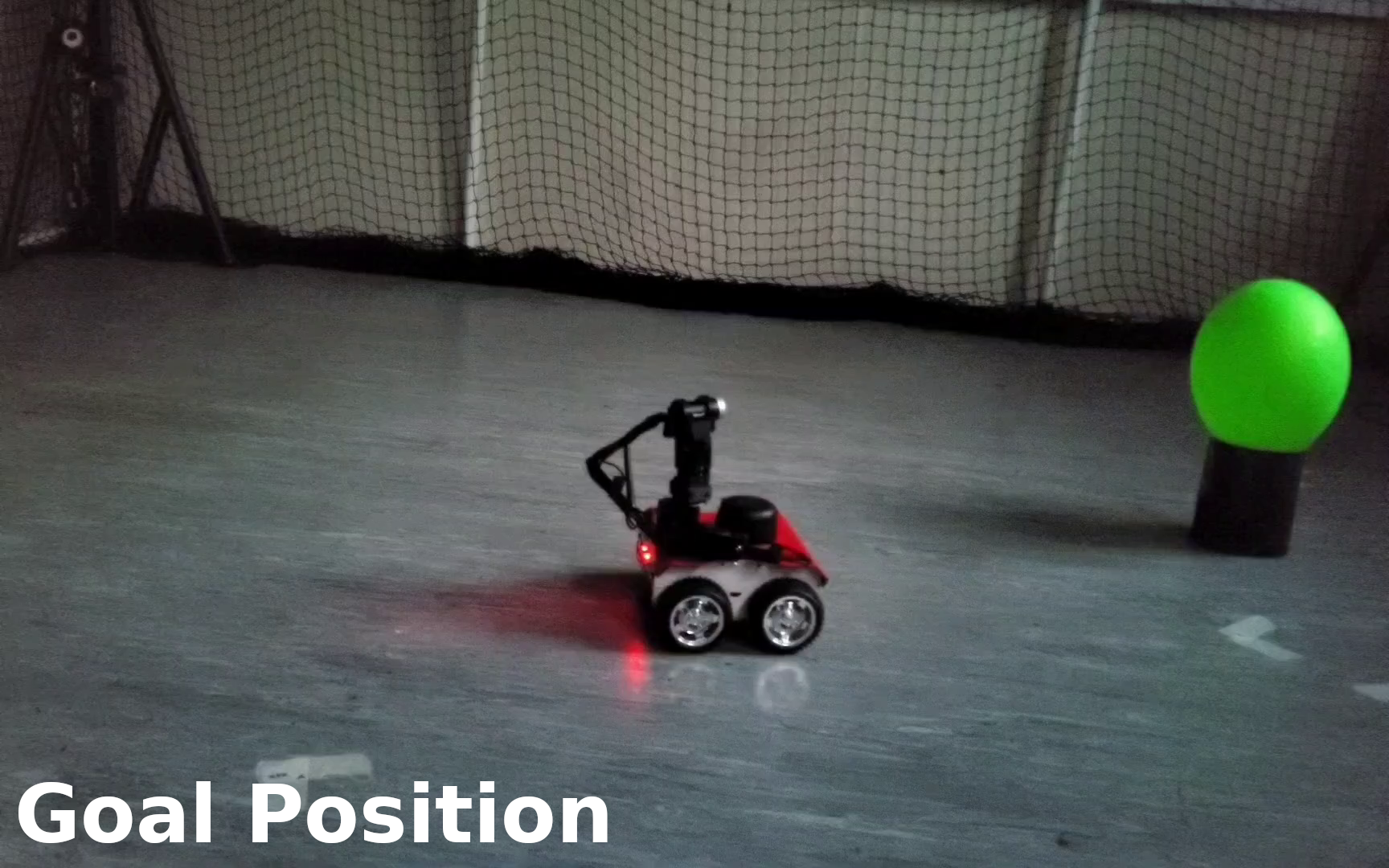}
\caption{Sequence of the dolly-in shot in the real world, showing the RL agent guiding our ground-based filming robot during the dolly-in shot from start to finish, using precise throttle, steering, pan, and tilt adjustments.}
\label{fig:real_world_experiment}
\end{figure*}


\section{Formalisation of dolly-in shot }
\label{sec:formalisation_of_dolly_in}

The dolly-in shot, a prevalent technique in filmmaking, emphasises the subject within a scene. Consider the subject as a simple green ball. The camera moves smoothly toward this subject \cite{Manetta1995, Nguyen2013}, with it centred in the frame, and concludes the shot once the subject occupies the desired portion of the image.
To analyse and model this technique, we must define it precisely and quantifiably. Such a definition allows us to apply RL to refine the reward function systematically, focusing on measurable factors:

\begin{itemize}
    \item The subject's proportion within the frame.
    \item The subject's position within the frame.
    \item The subject's orientation relative to the camera's movement.
\end{itemize}

Quantifying these elements enables the systematic optimisation of the dolly-in shot for cinematic impact.


\subsection{Quantitative Definition of Dolly-In Shot}

\emph{Image Processing:} We begin by segmenting the image frame into two distinct regions: the subject and the background. For this, we use colour thresholding for our green ball subject, using OpenCV \cite{opencv4.5.0}. This process generates a binary mask $g(x,y)$, where pixels belonging to the subject are assigned a value of 1, while the background is set to 0. Any subsequent noise removal is carried out using OpenCV.

After segmentation, we calculate the subject's geometric moments, which are computationally efficient descriptors for the subject's shape, size, and position \cite{Kotoulas2005ImageAU}, defined by:
    \begin{equation}
        M_{ij} = \sum_{x}\sum_{y}x^{i}y^{j} g(x,y)
    \label{def:moments}
    \end{equation}
These geometric moments are used to extract essential features of the subject, facilitating subsequent analysis.

\emph{Subject's Image Area:} The proportion of the frame occupied by the subject is quantified using the zeroth-order geometric moment, $M_{00}$, which represents the sum of all non-zero pixels within the binary mask of the subject. It is calculated as: \( M_{00} = \sum_{x}\sum_{y} g(x,y)\). To express this as a percentage of the frame, we calculate the area percentage \(A\) by dividing $M_{00}$ by the total number of pixels \(s\) in the image:
    \begin{equation}
        A = \frac{M_{00}}{s}
    \end{equation}

\emph{Position of the subject in Frame:} The subject's position within the frame is determined by its centroid, representing the average position of all constituent points, calculated using the first-order moments \(\mu_{pq}\):

\begin{equation}
\mu_{pq} = \sum_{x}\sum_{y}(x-\Bar{x})^{p}(y-\Bar{y})^{q} g(x,y)
\end{equation}
Where $\Bar{x}=\frac{M_{10}}{M_{00}}$ and $\Bar{y}=\frac{M_{01}}{M_{00}}$ are coordinates of the centroid. Here, \(M_{10}\) and \(M_{01}\) represent the sums of x and y coordinates of all non-zero pixels in the binary mask \(g(x,y)\).

\emph{Subject Offset Angle:} Determining the subject's offset angle relative to the robot's heading is crucial for executing smooth dolly-in shots and minimising counter-steering. This involves a two-step process: 

First, we calculate the camera offset angle $\alpha$, which indicates the subject's angular deviation from the centre of the camera's Field of View (FoV). This calculation uses the camera's fixed parameters: the pixel width \(w\), the FoV \(f\), the subjects' x-axis position \(O_p\) and the image's midpoint \(M = \frac{w}{2}\). The pixel offset $P$ is \(O_p - M\). The camera offset angle \( \alpha \) is calculated using the formula:
\begin{equation}
    \alpha = P \times \left(\frac{f}{2}/{M}\right)
\end{equation}
This formula adjusts the angular deviation based on the camera's FoV and the pixel width.

Next, we calculate the subject's offset angle orientation \(\theta\) by incorporating the pan-angle \( \theta_p \) from the robot's joint sensors. The overall offset angle \( \theta \), which guides the robot's alignment towards the subject, is then defined as:
\begin{equation}
    \theta = \alpha + \theta_p
    \label{eq:obj_offset_angle}
\end{equation}
This combined offset angle ensures the robot maintains a direct line towards the subject, facilitating a centred and steady approach in dolly-in shots.

\emph{Quantifying Outcomes:} The success of the performed dolly-in shot is quantified by assessing progress towards predefined goals concerning the subject's area and position within the frame. This assessment employs a distance metric \( \delta(p_A, p_E, p_{\text{max}}) \), defined as follows:

\begin{equation}
    \delta(p_A, p_E, p_{\text{max}}) = \begin{cases}
    \frac{\Delta p}{p_E}, & \text{if } p_A < p_E \\
    \frac{\Delta p}{p_{E} - p_{\text{max}}}, & \text{if } p_A \geq p_E
    \end{cases}
    \label{eq:general_equation}
    \end{equation}

Where \( \Delta p = |p_A - p_E| \) denotes the absolute difference between the actual value $p_A$ and the desired parameter value $p_E$. This metric, acting as a normalised relative error, measures how close the current shot is to the desired cinematic goal. A value closer to 0 indicates proximity to the optimal state, as outlined in Reference \cite{Matignon2006}.

\section{Twin Delayed Deep Deterministic Policy Gradient}
\label{sec:RL_agents}

The problem can be formulated as a Markov Decision Process (MDP)\cite{sutton2018reinforcement}, defined as a tuple $(\mathcal{S}, \mathcal{A}, P, R, \gamma)$. At each timestep $t$, an agent in state $s_t \in \mathcal{S}$ chooses an action $a_t \in \mathcal{A}$. The agent then receives a reward $r_{t+1}$ and transitions to the next state $s_{t+1}$, according to the probability distribution $P(s_{t+1}|s_t, a_t)$. The reward $r_{t+1}$ is drawn from the reward distribution $R(s_t, a_t, s_{t+1})$. The agent's objective is to maximise the expected future cumulative reward $\mathbb{E}[G_t|s_t]$, where $G_t$ is the discounted return defined as $G_t=\sum^T_{k=t+1}\gamma^{k-t-1}R_k$ and $\gamma\in [0,1]$ is the discount factor. We can therefore define the state-action value function as $Q_\pi(s, a)=\mathbb{E}[G_t|s_t, a_t]$, where the policy $\pi(a, s)$ maps a state $s$ to a probability distribution over actions $a$.

To navigate the continuous action space inherent in our study, we implemented the Twin Delayed Deep Deterministic Policy Gradient (TD3) algorithm \cite{fujimoto2018addressing}. TD3 addresses the overestimation bias found in similar algorithms by the innovative use of two Q-functions $Q_{\phi 1}$ and $Q_{\phi 2}$, along with a mechanism for delayed policy updates. These features significantly enhance the stability and accuracy of value estimation, which is crucial for achieving precise control in complex cinematographic tasks.

TD3 introduces target policy smoothing and action noise clipping to mitigate the exploitation of erroneous estimations, a critical advancement over previous methods. By adding clipped noise to target actions, $a'(s')=\text{clip}(\mu_{\theta_{\text{target}}}(s') + \text{clip}(\epsilon, -c, c), a_{\text{low}}, a_{\text{high}})$, where $\epsilon \sim \mathcal{N}(0, \sigma)$, the algorithm ensures actions remain within feasible bounds. Additionally, TD3 employs the minimum of both Q-functions to update the target state-action value, fostering a more disciplined learning process.

The learning process trains the Q-functions by regressing to a target that reflects the discounted future return, while the policy $\mu_{\theta}$ is refined to maximize $Q_{\phi_1}$. To maintain stability and ensure effective convergence, target networks are periodically updated using Polyak averaging, which blends the old and new parameters to achieve a balanced update rate. 

By implementing the TD3 algorithm, our research addresses the intricate control demands of robotic cinematography, demonstrating the potential of reinforcement learning in facilitating sophisticated, real-world tasks. By elucidating the algorithm's structure and our application methodology, we underscore the versatility and robustness of TD3 in handling the nuanced requirements of our study's objectives. This adaptation demonstrates the practicality of our approach and paves the way for future exploration in the realm of autonomous robotic filming, where precise control and adaptability are paramount.

\subsection{Agent Configuration: Actions, State Variables, and Reward Shaping}
\label{sec:agent_configuration}
This section outlines the RL agent configurations deployed in our experiments, focusing on their action and state spaces. It then defines the reward functions.

We detail independent and combined control strategies for operating throttle and steering actions, noting key distinctions: independent employs two separate agents for platform management. In contrast, combined control consolidates this within a single agent. Furthermore, we introduce a complex agent designed to utilise all degrees of freedom —throttle, steering, pan, and tilt— to execute dolly-in shots on a ground-based filming platform efficiently.

\begin{table}[h]
  \renewcommand{\arraystretch}{1.5} 
  \centering
  \begin{tabular}{c@{\hspace{1.7em}}c|c@{\hspace{1.7em}}c}
    \hline
    \makecell{State \\ Variable} & \hspace{1em} Description \hspace{1em} & \makecell{State \\ Variable} & \hspace{1em} Description \hspace{1em} \\
    \hline
    $s_1$          & Object area                          & $s_6$          & Centroid y-error \\
    $s_2$          & Object area error                    & $s_7$          & Pan angle \\
    $s_3$          & Centroid x-position                  & $s_8$          & Tilt angle \\
    $s_4$          & Centroid x-error                     & $s_9$          & Object offset angle \\
    $s_5$          & Centroid y-position                  &                &                \\
    \hline
  \end{tabular}
  \caption{State variables and descriptions for the RL agent.}
  \label{tab:state-variables}
\end{table}

\subsubsection{Action spaces} As detailed in Table \ref{tab:action-variables}, the action spaces include throttle control ($a_1$),  steering control ($a_2$), camera pan ($a_3$), and tilt control ($a_4$). These actions span from [-1, 1], allowing scaling and application within the simulation environment. Table  \ref{tab:comparison} shows the distinct functional roles across different agents, highlighting their specialisation.

    \begin{table}[h]
      \renewcommand{\arraystretch}{1.5} 
      \centering
      \begin{tabular}{c@{\hspace{1.5em}}cc}
        \hline
        Action Variable & \hspace{1em} Description \hspace{1em}   \hspace{1em} \\
        \hline
        $a_1$ & Throttle control (move forward/backward) \\
        $a_2$ & Steering control (turn left/right) \\
        $a_3$ & Camera pan control (turn left/right) \\
        $a_4$ & Camera tilt control (look up/down) \\ 
        \hline
      \end{tabular}
      \caption{Action variables and descriptions for agents.}
      \label{tab:action-variables}
    \end{table}

\begin{table}[h]
  \renewcommand{\arraystretch}{1.5} 
  \centering
  \begin{tabularx}{\linewidth}{c*{4}{>{\centering\arraybackslash}X}}
    \hline
    Agent & Throttle $a_1$ & Steer\newline$a_2$ & Pan\newline$a_3$ & Tilt\newline$a_4$ \\
    \hline
    Throttle agent & \checkmark & & & \\
    Steering agent & & \checkmark & & \\
    Combined agent & \checkmark & \checkmark & & \\
    Complex agent & \checkmark & \checkmark & \checkmark & \checkmark \\
    \hline
  \end{tabularx}
  \caption{Comparison of actions used by different agents.}
  \label{tab:comparison}
\end{table}

\subsubsection{State spaces} Table \ref{tab:state-variables} specifies the state variables used for all agents, tailored for their specific control tasks:

\begin{itemize}

    \item \emph{Independent Agents:} Function with unique state spaces. The Throttle agent's state space, \(\mathcal{S}_T = \{s_1, s_2\}\), contrasts with the Steering agent's, \(\mathcal{S}_S = \{s_3, s_4\}\), reflecting their specialised control focus.
    
    \item \emph{Combined Agent:} Utilises a comprehensive state space, \(\mathcal{S}_C = \{s_1, s_2, s_3, s_4\}\), amalgamating throttle and steering controls for holistic task management.
    
    \item \emph{Complex Agent:} Incorporates all available state and action variables for comprehensive platform control.
    
\end{itemize}

These configurations, with agents tailored to specific filming objectives, are key to controlling and understanding cinematographic tasks in this study.

\subsubsection{Reward Shaping}
To guide the agents towards an optimal policy, using the criteria detailed in Section \ref{sec:formalisation_of_dolly_in}, we utilise a reward function formulated as a weighted sum of \( n \) sub-objectives: \( R(s, a) = \sum_{i=1}^n w_i \cdot r_i(s, a) \). This structure tailors the reward strategy to the specific objectives of each agent, directly influencing their actions to achieve the desired outcomes. Each weight \( w_i \) signifies the importance of the corresponding sub-objective \cite{kusari_predicting_2020}. Each objective's reward is based on Equation \ref{eq:general_equation}, customised for the control objective unless noted differently.

For independent agents, reward functions focus on individual control tasks:
\begin{enumerate}[itemsep=0pt, topsep=0pt, partopsep=0pt]
    \item \emph{Throttle Control Agent:} Focuses on maintaining the subject's desired size within the frame. The reward, $ R_{\text{area, original}}(s, a)$, depends on the actual area \(a_A\), expected area \(a_E\), and the maximum permissible area \(a_{\text{max}}\), defined as:
    \begin{equation}
        R_{\text{area, original}}(s, a) = \delta(a_A, a_E, a_{\text{max}})
        \label{eq:area_reward}
    \end{equation}
    \item \emph{Steering Control Agent:} Aims to keep the subject centred within the frame. Its reward, $ R_{\text{position}}(s, a)$, is calculated based on the actual position \(p_A\), expected position \(p_E\), and the maximum allowed position \(p_{\text{max}}\):

    \begin{equation}
        R_{\text{position}}(s, a) = \delta(p_A, p_E, p_{\text{max}})
        \label{eq:position_reward}
    \end{equation}
\end{enumerate}

The Combined agent's reward function combines the area and position rewards, balancing them through weights $w_1$ and $w_2$:

\emph{Combined Agent:} The reward function is a weighted sum of the rewards using Equations \ref{eq:area_reward} and \ref{eq:position_reward}:
    \begin{equation}
        R_{\text{combined}}(s, a) = w_1 \cdot R_{\text{area, scaled}} + w_2 \cdot R_{\text{position}}
        \label{eq:reward_centralized}
    \end{equation}
Here, \(R_{\text{area, scaled}}\) is an adaptation of \(R_{\text{area, original}}\) to promote optimisation of the subject area, especially during smaller adjustments:
\begin{align*}
R_{\text{area, scaled}} = 
\begin{cases} 
-0.5 + \left( \frac{|p - k|}{k} \right) \times (-0.5) & \text{if } p \leq k \\
\frac{p - a_E}{a_E} \times 0.5 & \text{if } p > k 
\end{cases}
\end{align*}

In this context, \(k\) represents the threshold value of \(R_{\text{area, original}}\), and \(a_E\) is the target area.  

\emph{Complex Agent:} The reward function for the complex task, $R_{\text{complex}}(s, a)$, combines previously defined reward components with additional penalties for object offset and excessive action variations between consecutive time steps:

\begin{equation}
\begin{aligned}
    R_{\text{complex}}(s, a) &= w_1 \cdot R_{\text{area, scaled}} \\
    &+ w_2 \cdot R_{\text{position}} \\
    &+ w_3 \cdot R_{\text{object offset}} \\
    &+ \text{p}
\end{aligned}
\label{eq:reward_complex}
\end{equation}

This reward function includes components, each tailored to a specific aspect of the filming task:

\( w_1 \cdot R_{\text{area, scaled}} \): Rewarding desired subject size in the frame

\( w_2 \cdot R_{\text{position}} \): Rewarding desired subject position (Eq. \ref{eq:position_reward}).

\( w_3 \cdot R_{\text{object offset}} \): Penalising robot heading deviations from the subject using $\theta$ from Eq. \ref{eq:obj_offset_angle}.

\(p\): Penalty for actions exceeding the threshold between consecutive timesteps.

\subsection{Implementation Details}
We outline the architectures and training parameters for our TD3 algorithm implementation. Our framework employs distinct networks for the actor and critic, undergoing training across episodes - each with 1500 timesteps. Key hyperparameters include a batch size of 128, a learning rate of 0.0005, a discount factor of 0.99, and a tau value for soft updates of 0.005. An experience replay buffer with a capacity of 10 million ensures diverse transition sampling.

Both the critic and actor networks consist of three fully connected layers. The initial layer, with 400 nodes, processes state and action inputs. This is followed by a 300-node layer, leading to an output layer that estimates action values for the critic and outputs actions for the actor, scaled by the maximum action value. ReLU activation functions facilitate non-linear learning across all layers except the output. The Adam is the optimiser for both networks and exploration noise is added during training to improve the exploration of the action space.

\section{Experimental Results}
\label{sec:Experimental_Results}
This section details the experimental setup and outcomes, validating the effectiveness of RL agents in performing autonomous dolly-in shot tasks.

\subsection{Simulation Setup}
\label{sec:simulation_Setup}
We chose Pybullet \cite{coumans2019} for its accurate physics and environmental modelling, Pytorch \cite{NEURIPS2019_9015} for its robust machine-learning capabilities, and the OpenAI Gym framework \cite{brockman2016openai} for its standardised interface for reinforcement learning tasks.

\subsection{Robot System Setup}
\label{sec:robot_system_setup}

\begin{figure}[thbp]
\centering

\begin{subfigure}{\linewidth}
    \centering
    \includegraphics[width=0.3\linewidth]{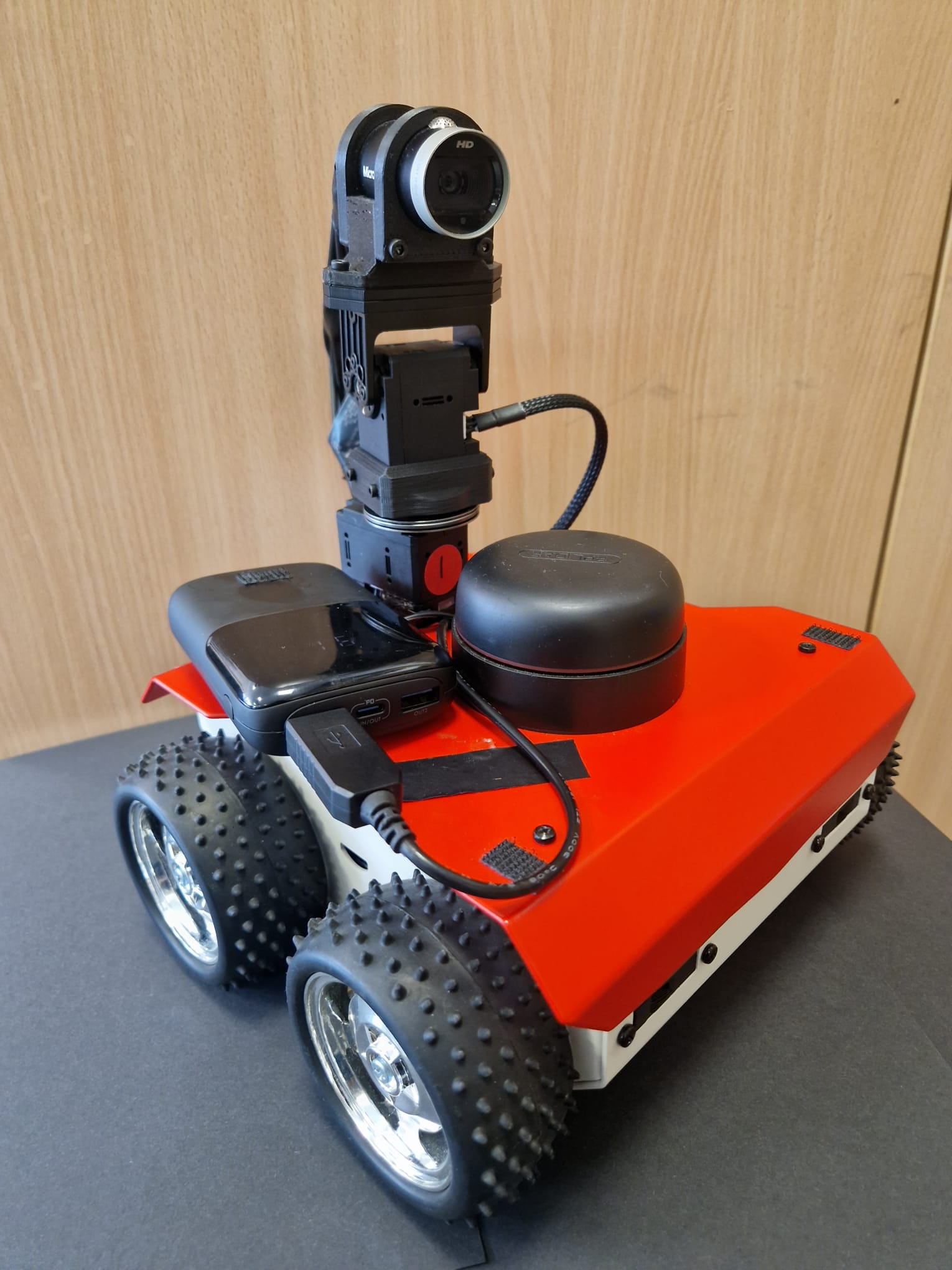}
    \includegraphics[width=0.53\linewidth]{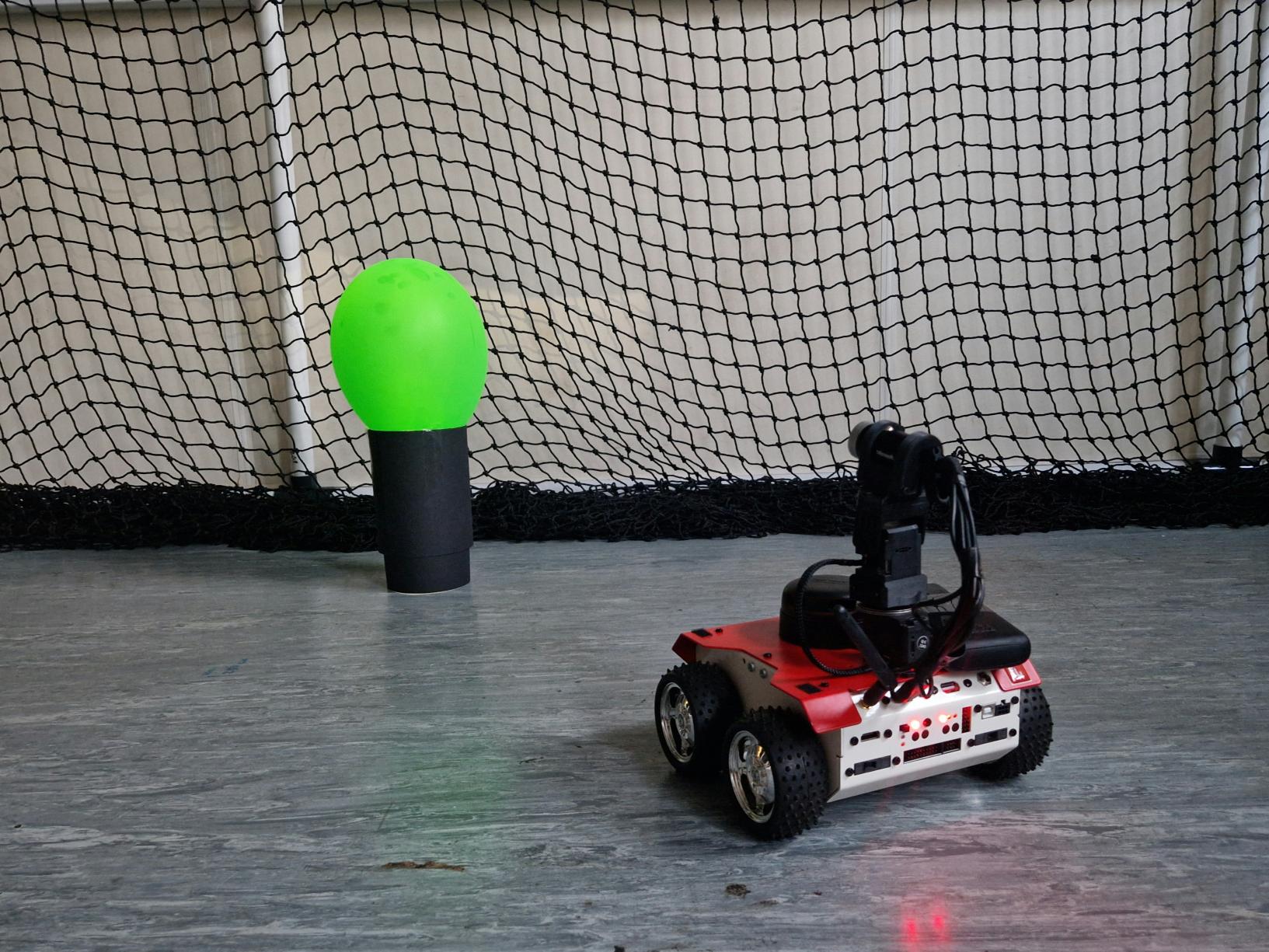}
        
    \caption{Modified ROSBot 2.0 Featuring Camera Turret for Filming Task.}
    \label{fig:modified_rosbot}
\end{subfigure}

\vspace{0.1cm} 

\begin{subfigure}{\linewidth}
    \centering
    \includegraphics[width=1.02\linewidth]{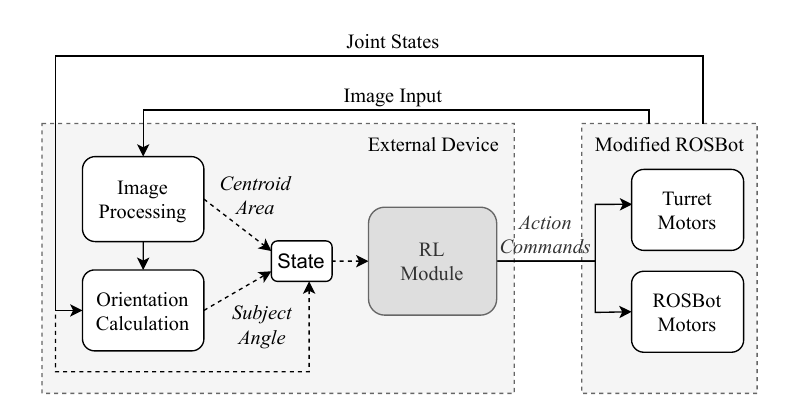}
    \caption{Data Flow and Software Component Diagram for the Modified ROSBot 2.0 in Filming Applications.}
    \label{fig:rosbot_software_diag}
\end{subfigure}

\caption{Modified ROSBot 2.0 for Filming: Physical Modifications Overview (Top) and Software Components with Data Flow Diagram (Bottom).}
\label{fig:system_rosbot}
\end{figure}

We customised a ROSBot 2.0 \cite{Husarion2020} for this study (Fig. \ref{fig:modified_rosbot}), integrating it with an Interbotix PhantomX Vision Tracking Robot Turret \cite{InterbotixRepo2022} and a Microsoft LifeCam Webcam for advanced tracking and filming capabilities. The ROSBot handles basic control and sensory functions. At the same time, an external computer powered by an Intel(R) Core(TM) i7-10875H CPU and Nvidia RTX 2080 Super GPU supports image processing and RL computations. We employ the Robot Operating System (ROS) \cite{ros} for efficient control and messaging. Figure \ref{fig:rosbot_software_diag} showcases the software architecture, highlighting the RL module and the system's information flow. After training in a simulated environment identical to our real world, the agent is ready for real-world validation to ensure its effectiveness outside the simulated conditions.

By bridging advanced hardware and software, this tailored robot system setup directly contributes to our research objectives of showcasing the practical applications of RL in robotic cinematography.

\subsection{Simulation Experiments}

Through simulation experiments, this section explores the efficacy of independent and combined control strategies for ground-based free-roaming robotic cinematography. Additionally, we assess an agent's performance with complete degrees of freedom against a baseline PD controller, focusing on the dolly-in shot task.

\subsubsection{Independent versus Combined}

This experiment assesses RL agents' utility, performance, and stability in developing optimal ground-based robotic cinematography systems, focusing specifically on the effectiveness of combined versus Independent control strategies. Our primary objective is to determine which strategy more effectively facilitates the execution of cinematography tasks, like the dolly-in shot, using throttle and steering actions with precision and stability.

To achieve this, we trained agents across 5,000 episodes, each lasting 1,500 timesteps. The agent setup, detailed in Section \ref{sec:agent_configuration}, contrasts two independent agents - a steering agent for subject alignment and a throttle agent for distance adjustment - with a combined agent tasked with collective goal achievement. This approach facilitates a nuanced comparison of their performance on the dolly-in task and their potential applicability to more complex challenges, such as incorporating pan and tilt mechanisms.

\begin{figure}[thbp]
  \centering
  
  \begin{subfigure}[b]{1\linewidth}
    \centering
    \includegraphics[width=0.87 \linewidth]{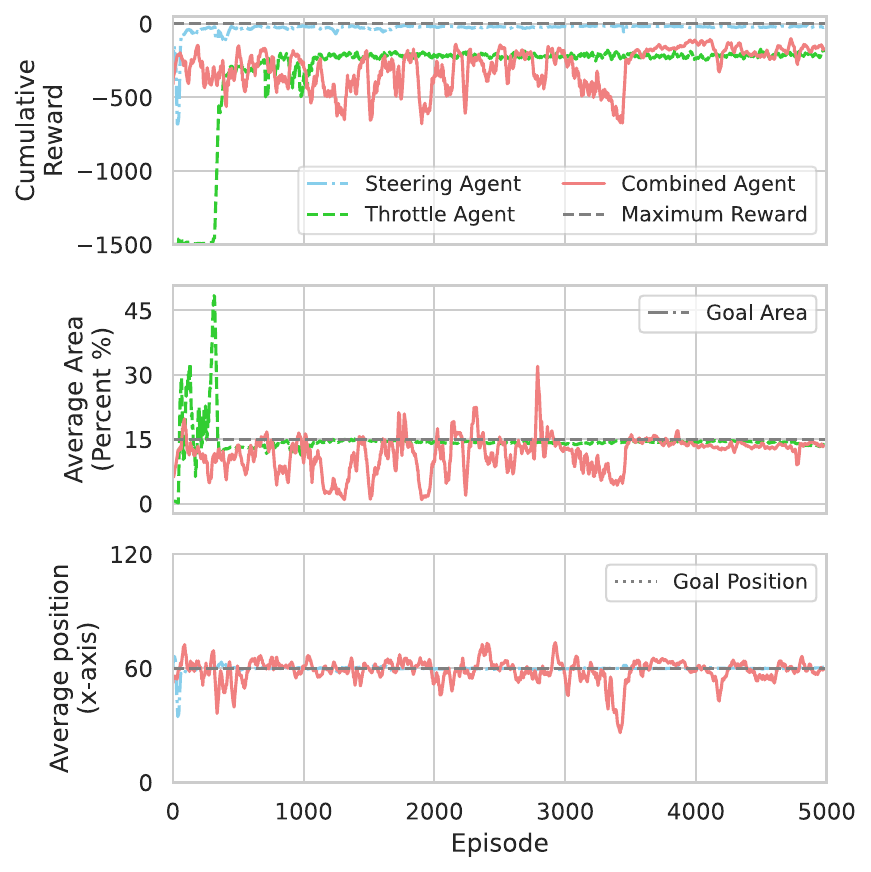}
    
    \caption{Training results for each agent during training. Cumulative Reward, Average Centroid Position (X-axis), and Average Area per episode.}
    \label{fig:decent_vs_cent}
  \end{subfigure}
  \hfill \\
  \begin{subfigure}[b]{1\linewidth}
    \centering
 
    \includegraphics[width=0.83\linewidth]{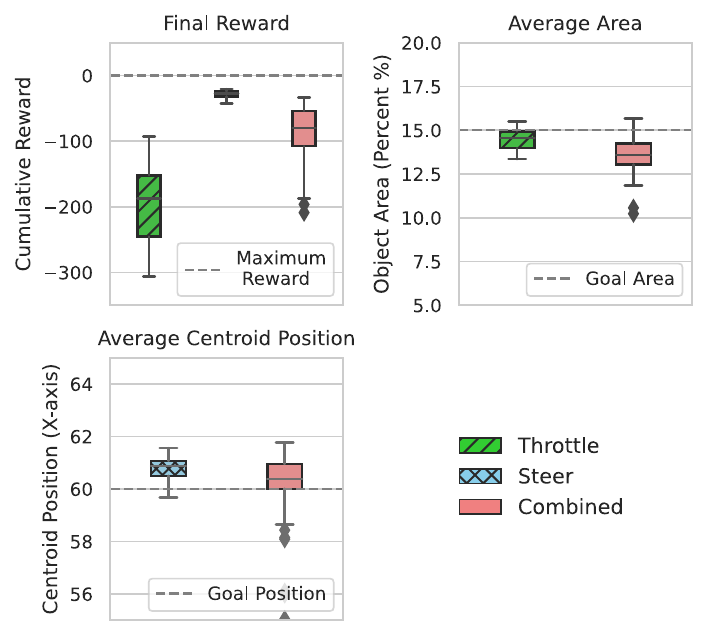}
    
    \caption{Stability evaluation box plot for each agent.}
    \label{fig:training_stability_eval}
  \end{subfigure}
  
  \caption{Training and stability evaluation for each agent.}
  \label{fig:combined_figure}
\end{figure}

\begin{figure}[thbp]
    \centering
  \includegraphics[width=0.83\linewidth]{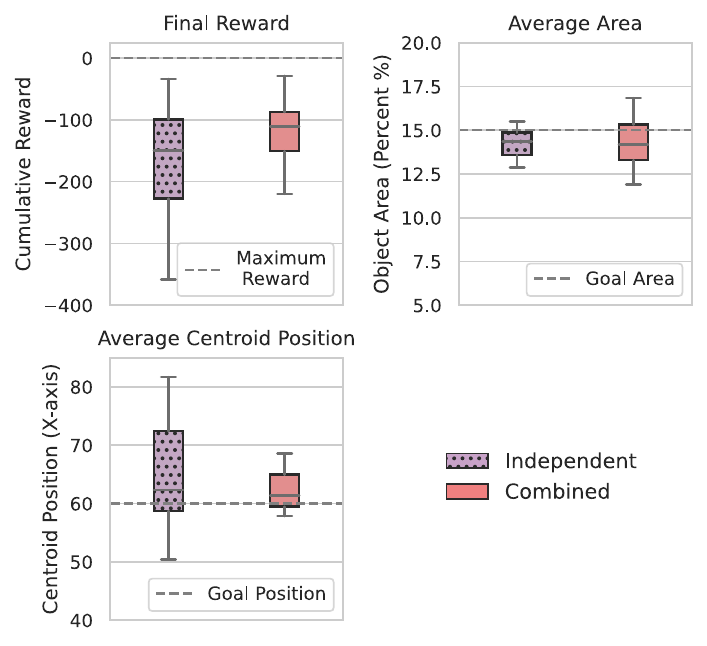}

\caption{Comparison of Combined and Independent control methods for dolly-in shot, over 100 trial runs.}
\label{fig:decent_vs_cent_global_eval}
\end{figure}

Our findings reveal distinct performance and stability patterns between the control strategies. Independent agents demonstrated quicker convergence (Fig. \ref{fig:decent_vs_cent}) and higher stability than their combined counterparts, as shown by a 100-trial stability test (Fig. \ref{fig:training_stability_eval}). However, when evaluating task performance over 100 trials using normalised reward metrics for area and position accuracy (Section \ref{sec:formalisation_of_dolly_in}), the combined strategy outperformed the independent approach, with a mean reward of -120 compared to -160. Scores closer to 0 indicate higher performance and efficacy, showing the centralised strategies performance in this context (Fig. \ref{fig:decent_vs_cent_global_eval}).

The performance variance, particularly in centroid accuracy, highlights coordination challenges inherent in decentralised setups. While the independent throttle agent often met area-size goals, steering accuracy was compromised, indicating a pressing need for cohesive control strategies. These results confirm the potential of our RL-based pipeline for simulating robotic cinematography tasks, setting a solid foundation for exploring basic dolly-in tasks and beyond.

\subsubsection{Full Dolly-in Task comparison}
Our analysis of the dolly-in task demonstrates that an RL-trained agent, utilising its full range of motion capabilities (throttle, steering, pan, and tilt), achieves performance comparable to a finely hand-tuned PD controller. This comparison underscores the RL agent's potential to match or surpass traditional control methods in robotic cinematography, marking a significant advancement in the field.

\begin{figure}[thbp]
  \centering
  
  \begin{subfigure}[b]{1\linewidth}
      \centering
      \includegraphics[width=0.86\linewidth]{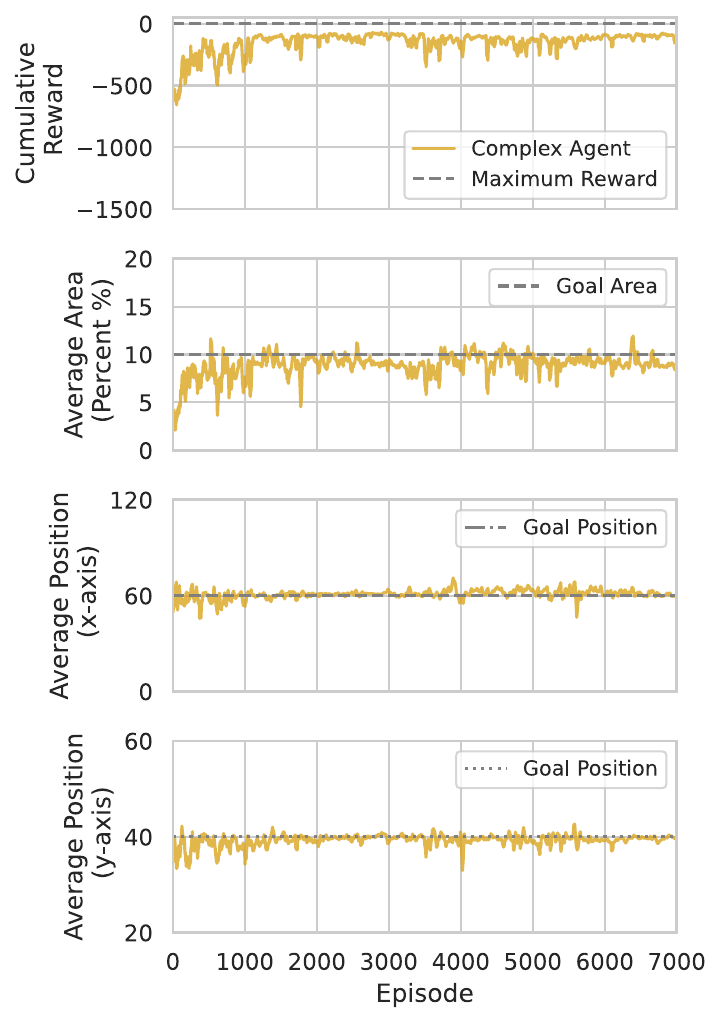}
      
      \caption{Training outcomes for the complex task agent: cumulative reward, average centroid position, and average area per episode.}
      \label{fig:complex_agent}
  \end{subfigure}
  \hfill \\
  \begin{subfigure}[b]{1\linewidth}
    \centering

    \includegraphics[width=0.83\linewidth]{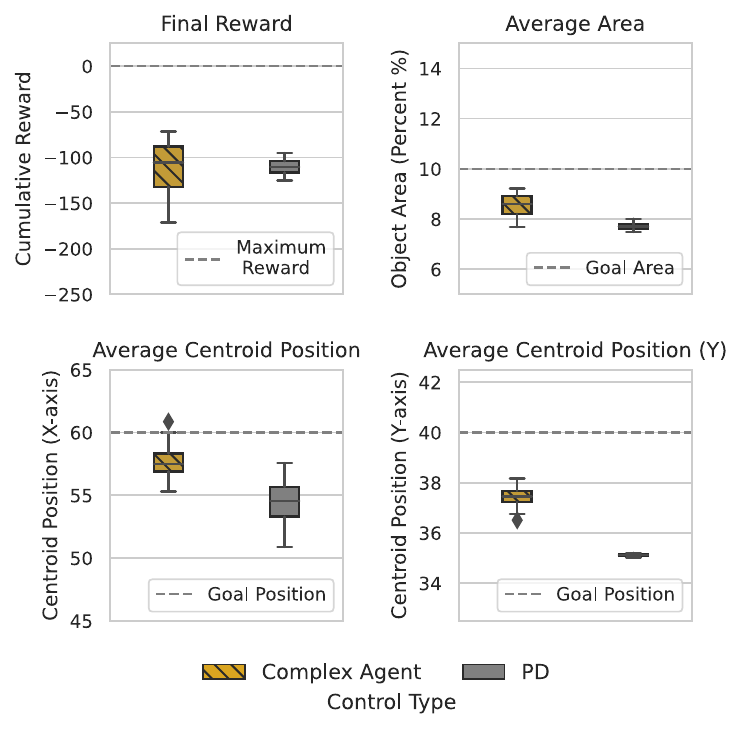}

\caption{Performance comparison between the RL agent and the baseline PD controller for dolly-in shots, using throttle, steering, pan, and tilt actions over 100 trials.}
\label{fig:final_sim_eval}

  \end{subfigure}
  
  \caption{Training and evaluation for complex agent.}
  \label{fig:complex_agent_figure}
\end{figure}

The complex agent was trained on the dolly-in task to conduct this comparison, incorporating strategies such as early stopping to prevent overfitting and ensure effective convergence. Progress in training the agent is detailed in Figure \ref{fig:complex_agent}. Concurrently, a PD controller was configured using the 'simple-pid' library (\cite{simplepid-1.2.0}), hand-tuned to achieve a performance goal aligned with that of the complex agent in a simulated environment.

The stability and performance of the complex agent and the classical PD controller were evaluated over 100 trials, as presented in Figure \ref{fig:final_sim_eval}. The results of this evaluation highlight the competitive performance of our trained RL agent compared to the PD controller. By closely matching the performance of traditional methods, our agent exhibits its capability to execute complex cinematographic tasks and signals a promising direction for the future of robotic cinematography.

\subsection{Real-world Experiments}

Bridging the 'reality gap' between simulation and real-world applications, our Sim2Real experiments aimed to validate the practical applicability of our RL model. Conducted in both simulated environments and real-world settings using the modified ROSBot 2.0 (Fig. \ref{fig:modified_rosbot}) and its digital twin, these experiments were designed to assess the generalisability and performance translation of our simulated agent into real-world scenarios.

Using the Simulation vs. Real-world Correlation Coefficient (SRCC, \cite{kadian2020}) as a quantifier of performance fidelity, where values closer to 1 indicate high simulation-to-real-world translation accuracy, our findings demonstrate the robustness and adaptability of the RL model. This validation marks a pivotal step in autonomous robotic filming, showing that our model can seamlessly transition from simulated training to effective real-world execution without intermediary fine-tuning.

In detail, the Sim2Real transfer experiment evaluated the agent's performance from three distinct starting positions (left, right, and centre) across both environments, with ten runs per position totalling 30 runs. The results, documented in Table \ref{tab:real2sim_experiment_SRCC}, reveal a consistently positive correlation between simulation and real-world performance, illustrating the agent's ability to achieve high rewards in both settings. This consistent performance underlines the model's capability for direct application in real-world cinematographic tasks, such as the dolly-in shot, further emphasising the potential of RL in bridging theoretical models with practical film-making applications (Fig. \ref{fig:real_world_experiment}).

Our comprehensive approach—from simulation training to zero-shot Sim2Real transfer—highlights a scalable pathway for integrating RL into robotic cinematography, advancing the field towards more autonomous and adaptable robotic filming solutions.

\begin{table*}[t]
    \centering
    
    \begin{tabular}{l*{4}{S[table-format=1.3, table-space-text-post=*, table-align-text-post=false, input-symbols=(), input-open-uncertainty=, input-close-uncertainty=, detect-all] *{3}{S[table-format=1.3, table-space-text-post=*, table-align-text-post=false, input-symbols=(), input-open-uncertainty=, input-close-uncertainty=, detect-all]}}}
    \toprule
    & \multicolumn{2}{c}{Cum. Reward} & \multicolumn{3}{c}{Object Area} & \multicolumn{3}{c}{Object Position (X-axis)} & \multicolumn{3}{c}{Object Position (Y-axis)} \\
    \cmidrule(lr){2-3} \cmidrule(lr){4-6} \cmidrule(lr){7-9} \cmidrule(lr){10-12}
    & {Sim.} & {Real.} & {Sim.} & {Real.} & {SRCC} & {Sim.} & {Real} & {SRCC} & {Sim.} & {Real} & {SRCC} \\
    \midrule
    Left & -170.77 & -176.27 & 10.12 & 9.63 & 0.69 & 59.42 & 61.22 & 0.52 & 44.0 & 44.54 & 0.65 \\
    Right & -166.03 & -166.35 & 10.09 & 10.38 & 0.80 & 59.32 & 59.08 & 0.72 & 44.1 & 44.23 & 0.83 \\
    Centre & -129.93 &-134.96 & 10.09 & 11.09 & 0.46 & 59.79 & 55.67 & 0.56 & 44.0 & 46.51 & 0.69 \\
    \bottomrule
    \end{tabular}

    \caption{Comparison of 30 runs (10 per starting position: left, right, centre) in both simulation and real-world settings, demonstrating Sim2Real transfer effectiveness with SRCC values indicating strong simulation-to-reality correlation.}
    \label{tab:real2sim_experiment_SRCC}
\end{table*}

\section{Discussion and Limitations}
\label{sec:discussion_lims}

Our research demonstrates the potential to use RL to automate traditional dolly-in shots with a ground-based, free-roaming filming robot. It highlights the synergy between robotics and data-driven filmmaking. This approach demonstrates RL's feasibility in complex cinematographic tasks and sets a precedent for future interdisciplinary innovations.

Our exploration of combined and independent control strategies reveals critical insights into their effectiveness and challenges. Combined control approaches offered enhanced action-objective synchronisation between actions and objectives but required longer training periods due to increased variables and larger search spaces. Despite these inherent challenges, combined control may be a viable alternative to independent systems, especially in complex, multi-action scenarios such as those found in ground-based robotic filming, which require a level of coordination similar to that in aerial cinematography. However, independent control and Multi-Agent Reinforcement Learning (MARL) remain promising for tasks with clear objective divisions, potentially mitigating coordination challenges and warranting further study.

Our findings demonstrate that our RL agent, trained for the complex task, can equal or even outperform traditional PD controllers in simulation, as shown in Figure \ref{fig:final_sim_eval}. The Sim2Real experiments critically validate the RL model's effectiveness for zero-shot transfer to real-world tests, showing a strong positive SRCC correlation. These findings underscore the utility of simulation-based training, demonstrating that performance in simulation can translate effectively to real-world applications, even on a limited, low-powered platform like the modified ROSBot 2.0. The TD3 algorithm, chosen for its strengths in continuous action spaces, proved practical for learning the dolly-in task. While this approach has proven effective, future work will explore comparisons with other RL methods to enhance performance further. However, it is important to note that there are currently no established benchmarks for this specific task, highlighting the novelty of our contribution.

Several challenges arose during real-world implementation, such as environmental variability and hardware limitations. The simplistic subject detection method required stable and consistent conditions, while the filming robot's underpowered equipment further constrained the task's complexity. These limitations, however, sharpened our focus on the core aspects of the problem. Through iterative testing and refinement of the RL model, we demonstrated its robustness and adaptability, which are crucial for real-world success.

The abstraction of the dolly-in shot task allowed us to isolate fundamental elements, setting the groundwork for effectively linking theoretical understanding with practical application, particularly in our real-world experiments. This approach led to the design of reward functions that emphasised key cinematic outcomes - subject size, position, and heading deviation - ensuring a platform-agnostic solution. As filming tasks and environments grow more complex, the reward function will need to be refined, introducing challenges in balancing reward design and component weightings.

With a more powerful robotic platform and a more robust and generalisable approach to subject detection, the need for high-fidelity digital twins becomes evident. While our approach is effective, it also provides ample opportunities for future development.

Imitation Learning (IL) offers a strategic approach to simplify reward design by utilising expert demonstrations to derive policies that capture desired filming behaviours. This approach could make the technology more accessible to non-experts. IL offers significant advantages over traditional RL by streamlining the learning process and more efficiently guiding the agent through the search space. This adaptability enhances the feasibility of RL-driven cinematography. It sets the stage for future research focused on expanding the applicability and robustness of our approach in dynamic and evolving filmmaking environments.

Our study illustrates the feasibility of RL in robotic cinematography for a ground-based filming platform and establishes a foundation for integrating theoretical understanding with practical filmmaking applications. The efficacy of combined strategies and the prospects of IL's in improving learning suggest new research and application opportunities. 

\section{Conclusions and Future work}
\label{sec:conclusions}

This study demonstrates the successful application of reinforcement learning (RL) to automate dolly-in shots with a ground-based filming robot, achieving performance comparable to traditional PD controllers. The effectiveness of the RL model, validated through simulation and real-world tests, represents a significant advancement in robotic cinematography. Combined control strategies showed considerable promise in synchronising actions and objectives for complex cinematic tasks, despite their complexity. This work lays a strong foundation for integrating RL into creative filmmaking, advancing the transition from theory to practice.

\subsection{Future Work}
To further refine the training process for cinematography agents, we plan to leverage imitation learning, utilising expert demonstrations to simplify and more effectively translate desired behaviours to RL agents. This approach aims to lower barriers for filmmakers, making advanced filmmaking technologies more accessible. Future research will focus on extending the applicability of RL in creative filmmaking, particularly in dynamic and evolving environments, thereby broadening the scope and impact of data-driven methods in the industry.

\bibliographystyle{ieeetr}
\bibliography{references}

\end{document}